\documentclass[a4paper]{cas-dc}

\usepackage[numbers,sort]{natbib}
\usepackage{enumitem}
\usepackage{placeins}
\usepackage{subcaption}
\usepackage{cleveref}
\usepackage{tikz}
\usetikzlibrary{arrows.meta,calc,backgrounds}
\usepackage{xparse}
\usepackage{graphicx}
\usepackage{booktabs}
\usepackage{xcolor}
\usepackage{etoolbox}
\usepackage{xspace}
\usepackage{microtype}
\usepackage[T1]{fontenc}
\usepackage[utf8]{inputenc}
\usepackage{amsmath,amssymb,amsfonts}
\usepackage{bm}
\setlist[itemize]{leftmargin=*} 
\setlist[enumerate]{leftmargin=*} 

\def\tsc#1{\csdef{#1}{\textsc{\lowercase{#1}}\xspace}}
\tsc{WGM}
\tsc{QE}

\begin{document}
\let\WriteBookmarks\relax

\shorttitle{One-Class Representation Learning for Rare Malignant Cell Detection}    

\shortauthors{Chatterjee et al.}  

\title[mode = title]{Needle in a Haystack: One-Class Representation Learning for Detecting Rare Malignant Cells in Computational Cytology}  

\author[1]{Swarnadip Chatterjee}
\cormark[1]
\ead{swarnadip.chatterjee@it.uu.se}

\author[2]{Vladimir Basic}

\author[3]{Arrigo Capitanio}

\author[1]{Orcun Goksel}

\author[1]{Joakim Lindblad}

\cortext[1]{Corresponding author}


\affiliation[1]{addressline={Department of Information Technology}, organization={Uppsala University}, country={Sweden}}

\affiliation[2]{addressline={Department of Clinical Diagnostics, School of Health Sciences}, organization={J\"{o}nk\"{o}ping University}, country={Sweden}}

\affiliation[3]{addressline={Department of Pathology}, organization={Link\"{o}ping University Hospital}, country={Sweden}}

\begin{abstract}
In computational cytology, detecting malignancy on whole-slide images is difficult because malignant cells are morphologically diverse yet vanishingly rare amid a vast background of normal cells. Accurate detection of these extremely rare malignant cells remains challenging due to large class imbalance and limited annotations. Conventional weakly supervised approaches, such as multiple instance learning (MIL), often fail to generalize at the instance level, especially when the fraction of malignant cells (witness rate) is exceedingly low. In this study, we explore the use of one-class representation learning techniques for detecting malignant cells in low-witness-rate scenarios. These methods are trained exclusively on slide-negative patches, without requiring any instance-level supervision. Specifically, we evaluate two one-class classification (OCC) approaches, Deep Support Vector Data Description (DSVDD) and Deep Representation One-class Classification (DROC), and compare them with fully supervised single-instance learning (FS-SIL), weakly supervised single-instance learning (WS-SIL), and the recent Iterative Self-Paced Supervised Contrastive Learning (ItS2CLR) method. The one-class methods learn compact representations of normality and detect deviations at test time. Experiments on a publicly available bone marrow cytomorphology dataset (TCIA) and an in-house oral cancer cytology dataset show that DSVDD achieves state-of-the-art performance in instance-level abnormality ranking, particularly in ultra-low witness-rate regimes ($\leq 1\%$) and, in some cases, even outperforming fully supervised learning, which is typically not a practical option in whole-slide cytology due to the infeasibility of exhaustive instance-level annotations. DROC is also competitive under extreme rarity, benefiting from distribution-augmented contrastive learning. These findings highlight one-class representation learning as a \textit{robust} and \textit{interpretable} superior choice to MIL for malignant cell detection under extreme rarity.
\end{abstract}


\maketitle

\section{Introduction}

Digital cytology \cite{kim2024digital_1,kim2024digital_2}, \emph{i.e.}, the high-resolution digitization of entire cytological slides, has become an important tool for early cancer screening, diagnosis, and longitudinal monitoring, enabling large-scale image analysis and AI-supported inspection of cellular morphology. Cytology is applied across multiple anatomical sites, including the oral cavity, bladder, cervix, breast, lung, and stomach, where diagnostic decisions often hinge on subtle changes in cell appearance and composition. The increasing availability of slide scanners, high-throughput acquisition systems, and cloud-based pathology archives has accelerated interest in computational cytology \cite{jiang2023deep,landau2019artificial} to support or automate parts of the diagnostic workflow to reduce workload, improve consistency, and aid earlier detection. However, reliable early malignancy detection remains very challenging, particularly when malignant cells are vanishingly rare within a vast background of normal cells.

A major difficulty in digital cytology is the extreme class imbalance intrinsic to real-world specimens, with normal cells overwhelmingly predominant even in slides that contain malignancy. Numerous studies across cervical, oral, pleural, and hematological cytology confirm that malignant or dysplastic cells are exceedingly rare, even in confirmed positive slides. A typical Pap smear slide may contain 100,000 to 300,000 squamous epithelial cells, but often only 10–20 malignant cells are present, resulting in witness rates (WRs) (the proportion of abnormal instances in positive samples) of 0.01\% or even lower \cite{bengtsson2014screening, cheng2021robust}. Median tumor cell loads of merely 0.23\% have been observed in malignant pleural effusions assessed by flow cytometry \cite{subira2023high}. In oral exfoliative cytology, neoplastic changes may need to be identified in oral smears that contain only a few abnormal cells \cite{carreras2015techniques}. In bone marrow aspirates, residual neoplastic cells can occur at minimal residual disease levels at around 0.05\%, corresponding to roughly one leukemic cell per 10,000 background cells, creating a similarly extreme class-imbalance setting \cite{kruse2020minimal,coustan2011new}.

Deep learning models, which are known to be datahungry and sensitive to class imbalance, often fail to generalize in these low WR regimes. Moreover, obtaining detailed cell-level annotations across an entire slide is prohibitively time-consuming and demands considerable cytological expertise (still only providing limited and subjective accuracy), rendering strongly supervised pipelines infeasible at scale.

\begin{figure*}[!t]
    \centering
    \includegraphics[width=\textwidth]{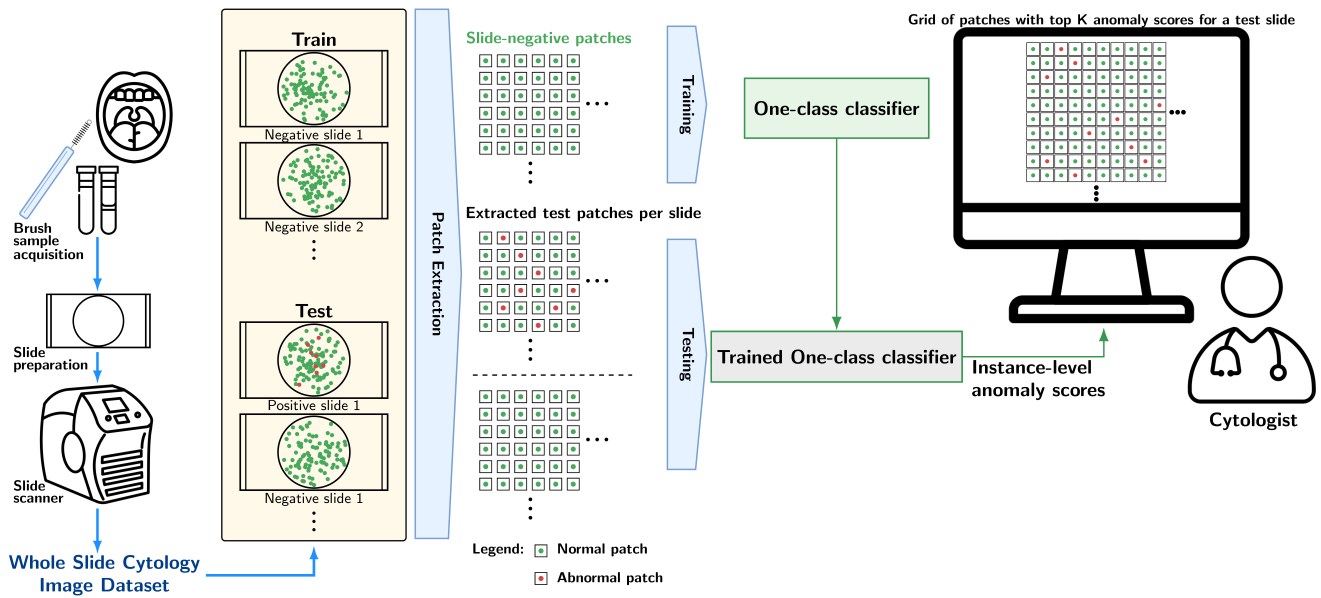}
    \caption{Overview of the proposed pipeline for rare abnormal cell detection in whole-slide cytology images. For training the one-class classifier, only slide-negative patches are used, which are known to be negative without requiring any instance-level annotations. At inference, patches from test slides are assigned anomaly scores, enabling the ranking and retrieval of rare malignant cells.}
    \label{fig:pipeline}
\end{figure*}
To circumvent these limitations, weakly supervised learning strategies such as Multiple Instance Learning (MIL) have gained traction in computational pathology \cite{campanella2019clinical}. In the MIL setting, a slide is treated as a bag of image patches (instances), and only the bag-level label (e.g., malignant or benign) is provided during training. Attention-based MIL (ABMIL) \cite{ilse2018attention} and clustering-constrained-attention multiple-instance learning (CLAM) \cite{lu2021clam,lu2021data} learn to identify key instances that drive the bag label, typically using attention mechanisms to weight the contribution of each patch. While these approaches have achieved success in histology and moderately imbalanced cytology tasks, they falter in the face of low WRs. When fewer than 1\% of patches are truly abnormal, attention mechanisms may converge on spurious patterns, leading to false positives or entirely missing the malignant cells.

These limitations motivate approaches that do not rely on positive instances during training. In response, one-class classification (OCC) has emerged as a promising alternative. OCC methods \cite{perera2020oca,ruff2021unifying} restrict training to only normal data and learn a compact representation of the normal class distribution. At inference, deviations from this normal manifold are flagged as anomalies. Deep Support Vector Data Description (DSVDD) \cite{ruff2018deep} exemplifies this idea by mapping normal instances into a latent feature space and constraining their embeddings within a hypersphere. The method is conceptually simple, robust to imbalance, and effective in domains where abnormalities are rare, diverse, and poorly annotated.

Complementing OCC, self-supervised learning (SSL) has transformed representation learning by removing the dependence on explicit labels. In particular, contrastive learning has gained prominence through frameworks such as SimCLR \cite{chen2020simple}, which learn embeddings by pulling together representations of two augmented views of the same instance while pushing them apart from representations of other instances in the batch—implemented via a contrastive objective whose denominator contrasts against many negatives. Recent work \cite{sohn2021learning} has connected SSL to OCC through Deep Representation One-class Classification (DROC), which learns self-supervised representations from one-class data and introduces distribution-augmented contrastive learning, using strong augmentations of normal samples to expand the training distribution and sharpen the notion of normality for anomaly detection at deployment.

In light of these methodological advances, our work aims to rigorously assess OCC approaches in real-world cytological settings, where the WR can drop below 0.1\%. Specifically, we adapt two well-established OCC frameworks, DSVDD and DROC and, evaluate on their ability to detect rare abnormal cells under simulated very-low witness-rate regimes of 0.05\% to 9\% (Fig.~\ref{fig:pipeline}). We perform experiments on two distinct whole slide cytology datasets: a publicly available bone marrow dataset~\cite{matek2021expert} and an in-house curated dataset of oral cancer Whole Slide Cytology Images (WSCIs). These datasets differ in specimen type and acquisition conditions, including stain protocols and color distributions, background debris, and slide-specific imaging artifacts, providing a challenging testbed for generalization.

To quantify how effectively each method identifies rare anomalies we compute top-$K$ instance-retrieval and ranking metrics and area under the FROC curve (AUFROC). We choose $K = 400$ for the performance metrics to provide a reasonably focused set of the most suspicious cells for cytopathologist review. We consider this a realistic clinical use scenario, in which the expert is presented with a distilled subset of 400 candidates from a whole-slide sample that may contain on the order of $10^5$ cells. Unlike prior works that focus on slide-level classification, our evaluation puts focus on the instance-level retrieval of abnormal patches. Accurate instance-level retrieval is an essential requirement for clinical interpretability and human-in-the-loop cytological assessment. In such a setting, experts can not only review high-scoring candidates but also refine the learned representations by confirming true abnormal cells or guiding the model through outlier exposure style feedback or other active learning approaches. This instance-first design provides a bottom-up path towards interpretable whole slide image analysis: per patch abnormality scores can be aggregated to derive reliable slide-level predictions, enabling a transparent and clinically viable approach that integrates algorithmic retrieval with expert-driven refinement and decision making.

This study builds on our previous work~\cite{chatterjee2024detection}, which was the first to introduce one-class classification for instance-level rare abnormal cell detection in whole-slide cytology, using DROC~\cite{sohn2021learning} for rare key instance detection. Here, we broaden the scope by considering controlled witness-rate regimes spanning 0.05\%–9\% and by benchmarking an additional OCC baseline, DSVDD \cite{ruff2018deep}, with task-specific adaptations, while keeping DROC consistent with our previous formulation. We further extend the evaluation to an oral cancer whole-slide cytology dataset to better assess robustness and generalization.

\vspace{2mm}

Our study makes the following key contributions:
\begin{itemize}
\item We show that in realistic low witness-rate regimes (1\% or below), weakly supervised MIL-style approaches degrade substantially for instance-level rare cell retrieval, whereas one-class approaches remain more robust and better recover rare key instances.
\item Across both a publicly available bone marrow cytology dataset and an in-house oral cancer whole-slide cytology dataset, DSVDD emerges as the strongest and most consistent one-class method, particularly in the low witness-rate regime that is most relevant for practical screening.
\item We validate our findings with blinded expert review by two cytology experts on the oral cancer dataset, demonstrating the practical relevance of top-ranked candidate retrieval for expert-facing analysis.
\end{itemize}

In doing so, we aim to bridge the gap between algorithmic development in unsupervised learning and the practical demands of cytopathological diagnostics. Our findings suggest that OCC methods, particularly when equipped with strong representation learning, hold substantial promise for tackling the challenge of rare malignant cell detection in whole slide cytology.

\section{Related Work}

The analysis of WSCIs poses unique challenges due to the extreme class imbalance, heterogeneity, and noise often present in such data. Traditional supervised learning methods rely on large-scale, densely annotated datasets, which is infeasible in cytology due to the cost and expertise required for instance-level labeling. Consequently, a range of alternative learning paradigms have emerged, including weakly supervised learning through multiple instance learning (MIL), self-supervised learning (SSL), and positive-unlabeled (PU) learning, which leverages a limited set of positive examples alongside abundant unlabeled data. 

When detailed and reliable (instance-level) annotations are feasible, fully supervised single-instance learning (FS-SIL) where a convolutional or transformer network is trained directly on densely annotated patches, remains a strong baseline in computational pathology.

\subsection{MIL for Whole Slide Image Analysis}

MIL has become a widely adopted weakly supervised approach for whole-slide image (WSI) analysis in computational pathology. The central premise of MIL is that only bag-level (slide-level) labels are available, and the model must infer instance-level evidence while producing an accurate bag-level prediction. A WSI is treated as a bag composed of multiple small image patches (instances), and a positive bag indicates the presence of at least one abnormal patch.

Early MIL approaches employed straightforward aggregation schemes such as max-pooling or mean-pooling across instance-level predictions. While computationally efficient, these methods lacked the capacity to distinguish truly informative patches. The introduction of ABMIL \cite{ilse2018attention} brought greater interpretability by assigning learnable attention weights to each instance, effectively highlighting potentially discriminative WSI regions. Subsequent work, such as CLAM \cite{lu2021clam}, incorporated clustering constraints to encourage feature diversity among attended patches, improving the model’s ability to localize abnormal regions and learn robust representations. Dual-stream MIL (DSMIL) \cite{li2021dual} further advanced MIL by introducing dual-stream processing of both instance-level and bag-level embeddings, incorporating contrastive objectives to improve feature discrimination.
In histopathology, where connectivity and neighbourhoods are relevant, graph-based approaches~\cite{pati2023weakly} have also been explored for WSI analysis. A closely related but simpler family of methods are weakly supervised single-instance learning (WS-SIL)~\cite{koriakina2024deep, lu2020deep} approaches, where all patches extracted from a tumor slide inherit the positive slide label (and patches from slide-negative cases inherit the negative label), and a standard patch classifier is trained on these noisy labels. While WS-SIL is easy to implement and
computationally convenient, it tends to overfit to slide-level biases and struggles to localize the truly abnormal cells when witness rates are very low. More advanced weakly supervised frameworks combine MIL with contrastive
representation learning. For example, Iterative Self-paced Supervised Contrastive Learning (ItS2CLR) \cite{liu2023multiple} iteratively refines pseudo-labels by selecting high-confidence instances using the aggregator’s instance-confidence scores and then optimizes a supervised contrastive objective on these pseudo-labels to highlight key instances within positive bags. Such methods aim to bridge the gap between coarse slide-level supervision and fine-grained instance localization, but their performance still degrades when malignant cells constitute only a tiny fraction of all instances, as is typical in WSCIs. In~\cite{pati2023weakly}, explainability scores of entity-based graph neural networks trained with WSI labels are used to generate instance-level weak labels.
Such spatial connectivity information, however, is less relevant in cytology.

Despite the above-mentioned innovations, MIL frameworks face a severe limitation in low WR settings. In cytology, where abnormal cells may represent less than 0.1\% of all instances, slide-level supervision becomes extremely weak. In such scenarios, MIL models often fail to localize the few critical abnormal instances, instead they focus on spurious features or irrelevant patterns. Label noise resulting from assigning uniform labels to all instances within positive bags can further hinder performance. These limitations motivate the exploration of alternative approaches that can operate effectively without relying on bag-level supervision or assumptions about instance distributions.

\subsection{SSL \& SSL-OCC Approaches}

\vspace{2mm}

SSL learns transferable representations from unlabeled data via surrogate training objectives, reducing reliance on costly manual annotations. A prominent family is \emph{contrastive learning}, where representations of two augmented views of the same patch are pulled together while other patches are pushed apart. SimCLR~\cite{chen2020simple} popularized this instance-discrimination formulation, MoCo~\cite{he2020moco} introduced a momentum encoder with a queue of negatives, BYOL~\cite{grill2020byol} removed explicit negatives via bootstrap prediction, and SwAV~\cite{caron2020swav} leveraged prototype assignments to enforce view-consistent cluster predictions. In WSI analysis \cite{schirris2022deepsmile, torpey2024deepset}, SSL is commonly used to pretrain patch encoders on large collections of unlabeled patches, after which the learned embeddings can be used by weakly supervised aggregators for slide-level prediction and key instance retrieval.

In cytology WSIs, where malignant cells are rare and appearance varies with staining differences and artifacts, SSL provides robust patch embeddings from abundant unlabeled data. Extensions to abnormality detection further use strong augmentations of normal samples to generate \emph{pseudo-anomalies}; DROC~\cite{sohn2021learning} proposed this distribution-augmented contrastive strategy, tightening normal representations and improving anomaly separability at inference. Although DROC was not originally trained or evaluated on medical images, our work adapts and evaluates it under realistic, cytology-relevant WR constraints to clarify its strengths and limitations in practical whole-slide cytology analysis.

\subsection{OCC in Medical Image Computing}

OCC methods \cite{fernando2021deep,tschuchnig2021anomaly,wei2018anomaly} in medical image computing have emerged as a robust alternative for settings with extreme class imbalance or unavailable anomaly annotations. In OCC, the model is trained exclusively on normal samples and aims to learn a compact representation of the normal class. Deviations from this representation at test time are considered anomalies.

A prominent method in this category is DSVDD~\cite{ruff2018deep}. DSVDD learns to embed normal samples into a latent space such that all embeddings lie within a minimal enclosing hypersphere. At inference time, samples falling far outside this hypersphere are identified as abnormal. DSVDD is particularly advantageous for cytology because obtaining a representative collection of abnormal cells across subtypes, staining conditions, and artifacts is challenging, while normal cells are abundant. This makes one-class methods especially suitable for scenarios in which malignant cells are extremely rare, sometimes accounting for far less than 1\% of the cells in a whole-slide specimen. A recent study by Marta et al.~\cite{marta2025anomaly} also explored anomaly detection for cytology images using benign-only training data. However, their objective differs substantially from ours: although patch-level anomalies are detected, the final task and evaluation are performed at the whole slide level, classifying a slide as malignant if at least one patch is anomalous, rather than retrieving rare abnormal cells at the instance level. In contrast, our work focuses on instance-level rare cell retrieval in whole-slide cytology under controlled low witness-rate settings, with ranking-based evaluation and expert review of top-ranked candidates.

\section{Methods}
\label{sec:method}
We introduce the use of one-class classifier based representation learning for
anomaly detection in whole slide cytology images (WSCIs); an overview of the pipeline is shown in Fig.~\ref{fig:pipeline}.

Unlike fully supervised or multiple-instance learning approaches, these frameworks are trained solely on slide-negative patches and identify abnormalities as deviations from the learned class manifold.

We evaluate five methods in total, including two state-of-the-art one-class representation learning approaches: \textbf{DSVDD}~\cite{ruff2018deep} and \textbf{DROC}~\cite{sohn2021learning}, both used with task-specific adaptations. Both methods are trained exclusively on slide-negative patches and aim to assign higher anomaly scores to abnormal cells at test time. For comparison, we also include three reference methods: (i) \textbf{FS-SIL} \cite{koriakina2024deep,lu2020deep}, a fully supervised single-instance learning model trained on patch-level labels, (ii) \textbf{WS-SIL} \cite{koriakina2024deep,lu2020deep}, a weakly supervised single-instance learning baseline based on slide-to-patch label inheritance, and (iii) \textbf{ItS2CLR} \cite{liu2023multiple} updates pseudo-labels using the aggregator’s instance-confidence scores and optimizes a supervised contrastive loss to emphasize key instances within positive bags. Together, these five methods span fully supervised, weakly supervised, and one-class anomaly detection paradigms.

\subsection{DSVDD}

Deep SVDD is a natural fit for our setting because it learns a compact representation of slide-negative (normal) cell-sized patches without requiring any abnormal training labels. 

Classical DSVDD \cite{ruff2018deep} learns a feature map
$\varphi \colon \mathbb{R}^{H\times W\times C}\rightarrow\mathbb{R}^{d}$
that embeds normal inputs $x_{i}$ into a latent space where they are
concentrated around a single center $c\in\mathbb{R}^{d}$.
Given network parameters $\theta$, the standard DSVDD objective is

\smallskip
\begin{equation}
\mathcal{L}_{\text{DSVDD}}(\theta)
    = \frac{1}{n} \sum_{i=1}^{n} \lVert \varphi(x_{i};\theta) - c \rVert_{2}^{2}.
\end{equation}
\smallskip

\noindent with the anomaly score for a test instance $x$ defined as

\smallskip
\begin{equation}
s_{\text{DSVDD}}(x)
    = \lVert \varphi(x;\theta) - c \rVert_{2}^{2}.
\end{equation}
\smallskip

\noindent The center is typically initialized as the mean of the initial features,

\smallskip
\begin{equation}
c = \frac{1}{n} \sum_{i=1}^{n} \varphi(x_{i};\theta_{0}),
\end{equation}
\smallskip

\noindent and kept fixed during optimization, while weight decay regularization is applied to the network parameters to discourage trivial solutions.

\medskip
\noindent\textbf{Autoencoder pretraining and center stabilisation.}
In our setting, we first pretrain an autoencoder $(f_{\text{enc}}, f_{\text{dec}})$
on normal patches using a reconstruction loss

\smallskip
\begin{equation}
\mathcal{L}_{\text{AE}}(\theta_{\text{enc}}, \theta_{\text{dec}})
    = \frac{1}{n} \sum_{i=1}^{n}
      \Bigl\lVert f_{\text{dec}}\!\bigl(f_{\text{enc}}(a(x_{i})); \theta_{\text{dec}}\bigr) - a(x_{i}) \Bigr\rVert_{2}^{2},
\end{equation}
\smallskip

\noindent where $a(\cdot)$ denotes an image-augmentation operator and
$f_{\text{enc}}$ shares the same encoder architecture as the DSVDD network.
After pretraining, we transfer the encoder weights and use

\smallskip
\begin{equation}
\varphi(x;\theta) = f_{\text{enc}}(x;\theta_{\text{enc}})
\end{equation}
\smallskip

\noindent as the DSVDD feature map. \\

\vspace{2mm}

\noindent The DSVDD center is then estimated on deterministically preprocessed normals
$\tilde{x}_{i} = t(x_{i})$ using the pretrained encoder:

\smallskip
\begin{equation}
c = \frac{1}{n} \sum_{i=1}^{n} \varphi(\tilde{x}_{i};\theta_{0}),
\end{equation}
\smallskip

\noindent where $t(\cdot)$ applies resizing and tensor conversion but no random augmentation.
To avoid coordinates of $c$ collapsing near zero, we apply an elementwise clamp

\smallskip
\begin{equation}
c_{j} \leftarrow \operatorname{sign}(c_{j})\,\max\bigl(\lvert c_{j}\rvert, \varepsilon\bigr),
\qquad j = 1,\dots,d,
\end{equation}
\smallskip

\noindent with a small constant $\varepsilon>0$. This fixed, non-degenerate center is used
throughout DSVDD training.

\medskip
\noindent\textbf{DSVDD training objective.}
Given the fixed center $c$, we train the encoder on augmented normals $a(x_{i})$
using the DSVDD objective

\smallskip
\begin{equation}
\begin{aligned}
\mathcal{L}_{\text{DSVDD}}(\theta_{\text{enc}})
    &= \frac{1}{n} \sum_{i=1}^{n}
      \Bigl\lVert \varphi\bigl(a(x_i);\theta_{\text{enc}}\bigr) - c \Bigr\rVert_2^2 \\
    &\quad + \lambda \sum_{l} \bigl\lVert W^{(l)} \bigr\rVert_F^2 .
\end{aligned}
\end{equation}
\smallskip

\noindent where the second term is $\ell_{2}$ weight regularization on the encoder weights
$\{W^{(l)}\}_{l}$ with coefficient $\lambda$.
The center $c$ is not updated during this stage.

To make the method robust given the comparatively small normal training set (9,185 patches) in the bone-marrow experiments, we use multi-seed ensembling and mild test-time augmentation to reduce variance in the anomaly ranking. 
For the oral-cancer dataset, the much larger pool of slide-negative training patches ($\sim$640k) stabilizes representation learning, so we keep DSVDD deterministic and avoid the extra cost of multi-seed and multi-view inference.

\medskip
\noindent\textbf{Seed ensemble and mild test-time augmentation.}
To reduce variance, we train $S$ independent DSVDD models (seeds) with parameters
$\{(\theta_{\text{enc}}^{(s)}, c^{(s)})\}_{s=1}^{S}$, obtained by repeating the
pretraining, center estimation, and DSVDD training with different random
initializations and data shuffles.

We employ test-time augmentation \cite{shanmugam2021better,kimura2021understanding,cohen2023boosting}, where at test-time, each image $x$ is optionally evaluated under a small set of
deterministic views

\smallskip
\begin{equation}
\mathcal{V}(x) = \{v_{0}(x), v_{1}(x), \dots, v_{K-1}(x)\}.
\end{equation}
\smallskip

\noindent where $v_{0}$ is the original view and the remaining $v_{k}$ are mild transformations
(horizontal flip and small rotations). For seed $s$ we define the per-view distances

\smallskip
\begin{equation}
d_{k}^{(s)}(x)
    = \Bigl\lVert \varphi\bigl(v_{k}(x);\theta_{\text{enc}}^{(s)}\bigr) - c^{(s)} \Bigr\rVert_{2}^{2},
\qquad k=0,\dots,K-1.
\end{equation}
\smallskip

\noindent Rather than averaging over all views, we form a convex blend between the original view and the most suspicious view:

\smallskip
\begin{equation}
\tilde{s}^{(s)}(x)
    = d_{0}^{(s)}(x)
      + \lambda_{\text{blend}}\Bigl(\max_{k} d_{k}^{(s)}(x) - d_{0}^{(s)}(x)\Bigr) \,,
\end{equation}
\smallskip

\noindent where $\lambda_{\text{blend}}\in[0,1]$ controls how strongly we emphasize the most abnormal-looking transform.
Finally, the ensemble anomaly score aggregates across seeds by averaging:

\smallskip
\begin{equation}
s(x) = \frac{1}{S} \sum_{s=1}^{S} \tilde{s}^{(s)}(x).
\end{equation}
\smallskip

\noindent This procedure combines robust representation learning (via autoencoder pretraining and strong data augmentation) with a stabilised center, multiple random restarts, and a view-aware scoring rule that favors any transformation that reveals abnormal
morphology, while still anchoring scores to the original cell appearance.

\subsection{DROC}

We evaluate DROC~\cite{sohn2021learning}, a SimCLR-style contrastive strategy adapted for one-class anomaly detection. The method does not require abnormal training examples; instead,
it generates \emph{pseudo-abnormal} variants of normal instances via strong augmentations. In the distribution-augmented
formulation of DROC, these variants are treated as \emph{separate instances} and therefore enter the
contrastive denominator as additional negatives, encouraging compact normal representations while pushing apart strongly
distorted views.

Let $f(\cdot)$ be an image encoder and $g(\cdot)$ a projection head, and define the projected embedding
\begin{equation}
\phi(x)=\frac{g(f(x))}{\lVert g(f(x))\rVert_2}.
\end{equation}
Given an input $x_i$, we sample two stochastic augmentations $A$ and $A'$ to obtain two views
$x_i^{A}=A(x_i)$ and $x_i^{A'}=A'(x_i)$. Using a contrastive loss, the objective for anchor $x_i^{A}$ is
\begin{equation}
\mathcal{L}_{\text{clr}}
= - \log
\frac{\exp\!\left(\phi(x_i^{A})^\top \phi(x_i^{A'}) / \tau\right)}
{\left[
\begin{aligned}
&\exp\!\left(\phi(x_i^{A})^\top \phi(x_i^{A'}) / \tau\right) \\
&\quad + \sum\limits_{\substack{j=1\\ j\neq i}}^{n}
\exp\!\left(\phi(x_i^{A})^\top \phi(x_j^{A}) / \tau\right)
\end{aligned}
\right]},
\end{equation}
where $\tau$ is a temperature parameter and $n$ is the batch size.

\noindent\textbf{Distribution augmentation and pseudo-abnormals.}
To strengthen one-class separability, we augment the training distribution using a set of strong transformations
$\mathcal{A}_D$ that produce pseudo-abnormal variants. Concretely, for each $x_i$ we sample $D\sim\mathcal{A}_D$
and form a pseudo-abnormal view $x_i^{D}=D(x_i)$. Following~\cite{sohn2021learning}, these pseudo-abnormal views are
considered separate instances and are contrasted against the anchor by including them as additional negatives:
\begin{equation}
\mathcal{L}_{\text{DA}}
= - \log
\frac{\exp\!\left(\phi(x_i^{A})^\top \phi(x_i^{A'}) / \tau\right)}
{\left[
\begin{aligned}
&\exp\!\left(\phi(x_i^{A})^\top \phi(x_i^{A'}) / \tau\right) \\
&\quad + \sum\limits_{\substack{j=1\\ j\neq i}}^{n}
\exp\!\left(\phi(x_i^{A})^\top \phi(x_j^{A}) / \tau\right) \\
&\quad + \sum\limits_{j=1}^{n}
\exp\!\left(\phi(x_i^{A})^\top \phi(x_j^{D}) / \tau\right)
\end{aligned}
\right]}.
\end{equation}

\noindent In our cytology adaptation, $\mathcal{A}_D$ comprises:
\begin{itemize}
    \item \texttt{CenterCropResize}: \texttt{CenterCrop(height=180, width=180, p=1)},
    \item \texttt{ColorJitter} (\texttt{p=1}),
    \item \texttt{GridDistortion} (\texttt{p=1}), and
    \item \texttt{ElasticTransform} (\texttt{p=1}),
\end{itemize}
which generate pseudo-abnormal variants by perturbing morphology and appearance.

\noindent The final training objective is
\begin{equation}
\mathcal{L}_{\text{DROC}}
= \mathcal{L}_{\text{clr}}
+ \alpha\,\mathcal{L}_{\text{DA}},
\end{equation}
where $\alpha$ controls the contribution of distribution augmentation.

\medskip
\noindent\textbf{One-class detector on learned representations.}
After contrastive pretraining, we discard the projection head $g(\cdot)$ and extract representations $z_i = f(x_i)$ for all normal training instances. We then fit a one-class SVM with a radial basis function (RBF) kernel, using $\nu = 0.1$ and $\gamma = \texttt{auto}$, on $\{z_i\}$ to model the inlier distribution. At test time, the SVM decision function provides an anomaly score that we use to rank candidate abnormal cells.

\subsection{Reference Methods}

To contextualize the performance of one-class representation learning approaches, we compare them against the following baseline methods based on different levels of supervision:

\begin{itemize}
    \item \textbf{FS-SIL:} A fully supervised single-instance learning baseline trained using true instance-level labels (normal vs.\ abnormal), optimized with standard cross-entropy (optionally class-weighted). This method assumes exhaustive cell-level annotations and therefore serves as an upper-bound reference, although it is typically impractical in whole-slide cytology due to annotation cost and extreme abnormal-cell rarity.

    \item \textbf{WS-SIL via Label Inheritance:} A weakly supervised single-instance learning baseline~\cite{koriakina2024deep,lu2020deep} where each patch inherits its slide/bag label. Consequently, normal cells extracted from slide-positive specimens are treated as abnormal during training, yielding noisy instance labels. The model is trained with standard binary cross-entropy on these inherited labels. While commonly used as a simple baseline in WSI pipelines, its precision and generalization degrade substantially under low-WR regimes.

    \item \textbf{ItS2CLR:} Iterative Self-paced Supervised Contrastive Learning~\cite{liu2023multiple} is a weakly supervised contrastive framework for key-instance discovery under bag-level supervision. During an initial warm-up phase, negatives are drawn from slide-negative bags, while candidate positives are selected from slide-positive bags using the aggregator’s instance-confidence scores. After warm-up, ItS2CLR performs self-paced pseudo-labeling within slide-positive bags by selecting the top $r\%$ most confident instances as pseudo-positives and the bottom $r\%$ least confident instances as pseudo-negatives, with $r$ gradually increased over training in a self-paced setting. The method alternates between (i) updating these pseudo-label sets via confidence-based selection and (ii) refining the encoder using a supervised contrastive objective defined on the resulting pseudo-labels, progressively sharpening instance-level supervision across iterations.
\end{itemize}

\section{Datasets}

To evaluate the effectiveness and generalizability of our proposed framework, we conduct experiments on two cytology datasets: one having bone marrow smears \cite{matek2021expert} and another with oral liquid brush cytology slides. These datasets vary in terms of cell morphology, label granularity, and visual artifacts, providing complementary testbeds for benchmarking rare abnormal cell detection.

\subsection{Bone Marrow Cytomorphology Dataset}

The bone marrow dataset~\cite{matek2021expert,matek2021highly} contains high-resolution single-cell images extracted from digitized bone marrow smears. Each cell is annotated by expert hematologists into one of 21 morphological categories. In our study, we make a \emph{task-specific} choice to treat the Lymphocyte (LYT) category as the normal class, and we define an abnormal class by pooling seven minority categories: Abnormal eosinophil (ABE), Basophil (BAS), Faggott cell (FGC), Hairy cell (HAC), Smudge cell (KSC), Immature lymphocyte (LYI), and Other cell (OTH). This construction intentionally increases abnormal-class diversity (e.g., LYI reflects immature lymphocytes), providing a heterogeneous set of atypical morphologies for evaluating rare-cell detection.

For anomaly detection, only LYT cells are used to train the one-class models, while the pooled seven-class abnormal set is used exclusively for evaluation. We further vary the abnormal instance frequency to simulate different witness rates (WRs). Notably, this abnormal-class construction differs from our previous setup in~\cite{chatterjee2024detection}, which used a different class grouping and split configuration. Table~\ref{tab:pkgbm_cell_dist} summarizes the resulting cell-type counts and train-test split.
\vspace{0.5em}
\begin{table}[pos=h]
\caption{Cell-level statistics for the bone marrow dataset~\cite{matek2021expert}. All abnormal classes are merged into a single abnormal group for anomaly detection.}
\label{tab:pkgbm_cell_dist}
\centering
\small
\begin{tabular}{@{}lccc@{}}
\toprule
\textbf{Cell type} & \textbf{Total} & \textbf{Train} & \textbf{Test} \\
\midrule
LYT (normal)   & 26{,}242 & 18{,}369 & 7{,}873 \\
BAS (abnormal) & 441      & 308      & 133     \\
HAC (abnormal) & 409      & 286      & 123     \\
OTH (abnormal) & 294      & 205      & 89      \\
LYI (abnormal) & 65       & 45       & 20      \\
FGC (abnormal) & 47       & 32       & 15      \\
KSC (abnormal) & 42       & 29       & 13      \\
ABE (abnormal) & 8        & 5        & 3       \\
\bottomrule
\end{tabular}
\end{table}

\begin{table}[pos=h]
\caption{Number of nuclei per slide in each partition (sorted in decreasing order). Bold indicates the two slides shown later in Fig.~\ref{fig:oral_mosaic_a008} and Fig.~\ref{fig:oral_mosaic_a006}.}
\label{tab:nuclei_counts}
\centering
\begin{tabular}{@{}lp{.72\columnwidth}@{}}
\toprule
\textbf{Category} & \textbf{Number of patches (per slide)} \\
\midrule
Train Normal & 137\,646, 115\,382, 85\,149, 73\,573, 71\,033, 64\,455, 45\,155, 20\,016, 15\,901, 12\,795 \\
Train Tumor  & 64\,483, 37\,206, 34\,765, 33\,766, 30\,397, 26\,879, 16\,726, 7\,793, 5\,743, 4\,452 \\
Test Normal  & 75\,155, 73\,759, 44\,545, 28\,466 \\
Test Tumor   & 49\,534, \textbf{23\,257}, 15\,403, \textbf{9\,445} \\
\bottomrule
\end{tabular}
\end{table}

\subsection{Oral Cancer Cytology Dataset}

We also evaluate on an in-house oral cancer whole-slide cytology dataset \cite{koriakina2024deep, lu2020deep} comprising liquid-based cytology (LBC) brush specimens prepared as Papanicolaou (Pap) stained slides. All samples were collected at the Department of Orofacial Medicine, Folktandv\aa rden Stockholms l\"an AB, where for each patient a brush was used to scrape cells from clinically relevant areas of the oral cavity. Slide-level ground-truth labels (normal vs.\ malignant) were obtained via conventional gold-standard histopathological examinations.

The slides were digitized using a NanoZoomer S60 digital slide scanner (40$\times$, 0.75 NA objective) with multi-focus acquisition at 11 z-offsets ($\pm 2\,\mu$m, step size $0.4\,\mu$m), yielding RGB whole-slide images of size $103{,}936\times 107{,}520\times 3$ at $0.23\,\mu$m/pixel. 

For our study, we selected 28 labeled WSCIs: 20 for training (10 normal, 10 malignant) and 8 for testing (4 normal, 4 malignant). To obtain cell-level instances, we first detect nuclei locations and then extract cell-sized RGB patches of size $192\times192$ pixels centered at each detected nucleus, chosen to capture a single cell with most of its cytoplasm as well as the nucleus. Since each slide is available as a z-stack at the maximum magnification, we estimate focus across the focal levels and retain the best-focused patch per nucleus. Table~\ref{tab:nuclei_counts} summarizes the number of patches extracted from each slide across the training and testing partitions.


\section{Experimental Setup}

\subsection{Training and Evaluation Protocols}

We report experiments on two datasets with distinct supervision availability: (i) a bone marrow dataset\cite{matek2021expert} with patch-level labels, and (ii) an in-house oral cancer whole-slide cytology dataset with slide-level labels only. Below we describe the training and evaluation protocols used for each dataset and baselines.

\medskip
\noindent\textbf{Bone marrow dataset}
We use the 18{,}369 normal and 910 abnormal training images (Table \ref{tab:pkgbm_cell_dist}) to construct consistent supervision regimes for fully supervised, weakly supervised, and one-class methods. Fully supervised experiments (FS-SIL) are trained directly on these instance-level labels.

To ensure a controlled and comparable setup across weakly supervised and one-class baselines, we first partition the 18{,}369 training instances into 10 bags of equal size and keep this partition fixed for all such experiments (WS-SIL, ItS2CLR, DSVDD, and DROC). We then simulate different witness rates (WRs) by injecting controlled fractions of abnormal instances into 5 of the bags, while keeping the remaining 5 bags strictly normal (i.e., no injection). The one-class methods (DSVDD and DROC) are trained exclusively on these 5 normal bags, which together contain 9{,}185 images and are treated as the normal training set; the mixed bags are included along with the normal bags to support weakly supervised training (WS-SIL and ItS2CLR) under bag-level labels.

For evaluation, we use the full test set (7{,}873 normal and 396 abnormal instances) as a single evaluation pool (treat like one large merged bag). We compute anomaly scores (softmax scores or attention scores for FS-SIL, WS-SIL and ItS2CLR) for all instances and rank them in decreasing order. For FS-SIL, WS-SIL, DSVDD, and DROC, evaluation is performed purely in terms of instance ranking/retrieval within this pool (i.e., identifying highly suspicious cells). For ItS2CLR, the same pool is used to assess its instance selection and ranking performance under the bag-based formulation.

\medskip
\noindent\textbf{Oral cancer dataset}
For the oral cancer dataset, only slide-level ground truth is available; therefore, FS-SIL is not applicable. We use 28 slides in total: 20 slides for training (10 healthy, 10 malignant) and 8 slides for testing (4 healthy, 4 malignant). In all experiments on this dataset, we operate only with bag-level labels at the slide level.

For DSVDD and DROC, we train exclusively on patches extracted from the 10 healthy training slides. At test time, we apply the trained models to all patches extracted from each of the 8 test slides, producing per-patch anomaly scores that are ranked in decreasing order within each slide. To support expert review, we present the top-100 most anomalous patches per test slide to a cytologist and a cytotechnologist for qualitative assessment.

For WS-SIL, we train a patch classifier using slide-to-patch label inheritance at the slide level (healthy vs.\ malignant) and then rank all test patches by their predicted abnormal softmax scores, again extracting the top-100 per test slide for expert review. For ItS2CLR, we focus expert inspection on predicted-positive (malignant) test slides: we present the top-100 patches (based on the attention scores) for those test slides that are both slide-positive and predicted malignant by the model, reflecting the intended clinical use where attention is directed to slides flagged as suspicious.

\begin{table}[t]
\caption{Number of abnormal cells used for training and evaluation at each witness rate (WR).}
\label{tab:wr_cell_counts}
\centering
\small
\begin{tabular}{@{}ccc@{}}
\toprule
\textbf{WR (\%)} & \textbf{Train abnormal cells} & \textbf{Test abnormal cells} \\
\midrule
9    & 910 & 396 \\
5    & 455 & 198 \\
1    & 90  & 40  \\
0.5  & 45  & 20  \\
0.1  & 10  & 4   \\
0.05 & 5   & 2   \\
\bottomrule
\end{tabular}
\end{table}

\subsection{Bone marrow dataset: WR Variation}

We define the witness rate (WR) as the fraction of abnormal instances in the (weakly labeled) mixed training bags and evaluate performance across
$\{0.05, 0.1, 0.5, 1, 5, 9\}\%$. For each WR, we fix the corresponding number of abnormal instances used for training and for evaluation as summarized in Table~\ref{tab:wr_cell_counts}.

\noindent\textbf{Training WR control.}
Within a given WR setting, we distribute the WR-specific abnormal count across the mixed training bags such that no bag contains a substantially higher abnormal fraction than the others. This uniformity prevents an \textit{easier} bag from inadvertently revealing a stronger supervision signal and allows us to assess how well each method learns when supervision is consistently weak across all mixed bags.

\noindent\textbf{Test construction and merged evaluation.}
At evaluation time, we sample the WR-specific number of abnormal instances from the test pool (Table~\ref{tab:wr_cell_counts}), but we rank all instances on a single merged evaluation set. Concretely, the test set can be viewed as two bags: (i) a normal bag containing roughly half of $7{,}873$ normal instances, and (ii) a second bag containing the remaining half of $7{,}873$ normal instances plus the sampled abnormal instances (e.g., 396 for WR=9\%, and proportionally fewer at lower WRs). We do not evaluate these bags separately; instead, we merge them and compute a global instance ranking over all test instances. Because this merged pool contains a large fixed normal background, \emph{the effective abnormal prevalence in the ranked test set is typically lower than the nominal training WR, making evaluation intentionally more challenging while keeping the abnormal counts matched to the target WR}. To reduce variance at ultra-low WRs, we generate 10 random test trials per WR by resampling the abnormal subset and report results aggregated across these trials.

\noindent\textbf{Why WR variation is not performed for the oral cancer dataset.}
Unlike the bone marrow dataset, the oral cancer slides provide only slide-level ground truth (normal vs.\ malignant) and do not include exhaustive cell-level annotations identifying which specific patches correspond to malignant cells. Without knowing the abnormal instances explicitly within malignant slides, it is not possible to inject or subsample malignant cells in a controlled manner to realize targeted WR levels. Moreover, as a real-world clinical dataset, it is natural that different malignant slides (in both train and test) exhibit different, unknown witness rates, making controlled WR matching impractical and comparisons to the bone marrow WR-sweep setting inherently non-identical.

\subsection{Evaluation Metrics}

Our evaluation is motivated by a practical screening workflow: a whole-slide cytology image can contain on the order of $10^5$ detected cells (Table~\ref{tab:nuclei_counts}), making reliable exhaustive manual review infeasible. We therefore treat instance scoring primarily as a \emph{retrieval} problem, where the goal is to surface a small set of highly suspicious candidates for expert inspection. Below we describe the evaluation protocol for each dataset.

\medskip
\noindent\textbf{Bone marrow dataset (top-K retrieval).}
Let $\mathcal{T}=\{(x_i,y_i)\}_{i=1}^{N}$ denote the test instances with labels $y_i\in\{0,1\}$ (normal/abnormal), and let $\hat{s}_i$ be the predicted abnormality score for instance $x_i$. We rank all instances by $\hat{s}_i$ and evaluate performance in the top-$K$ retrieved set, with $K=400$.

Because presenting the retrieved candidates in ranked order can influence human reviewers toward the topmost positions, we use an unordered presentation for the qualitative expert evaluation in this paper, so that the assessment reflects only how well the model is able to surface suspicious cells within the top-$K$ set. Accordingly, we emphasize Recall@$K$ as our primary measure of effectiveness. In practical deployment, however, we envision the top-$K$ candidates being presented in ranked order according to the model score, allowing cytologists or pathologists to inspect the most suspicious cells first.
Using $\mathrm{TP}_K$ and $\mathrm{FN}_K$ computed within the top-$K$ list,
\begin{equation}
\mathrm{Recall@}K=\frac{\mathrm{TP}_K}{\mathrm{TP}_K+\mathrm{FN}_K}.
\end{equation}

\noindent As secondary analyses, we also report ranking-sensitive metrics that probe how early true abnormal cells appear among the highest-scoring candidates. Specifically, we compute \textbf{AUTK@}K (area under the top-$K$ curve),
\begin{equation}
\mathrm{AUTK@}K=\sum_{k=1}^{K}\frac{\mathrm{TP}_{\leq k}}{T},
\end{equation}
where $T$ is the total number of abnormal instances in the test set and $\mathrm{TP}_{\leq k}$ is the number of true positives within the top-$k$ ranked instances; \textbf{DCG@}K (discounted cumulative gain),
\begin{equation}
\mathrm{DCG@}K=\sum_{i=1}^{K}\frac{2^{y_i}-1}{\log_2(i+1)},
\end{equation}
and \textbf{nDCG@}K (normalized discounted cumulative gain),
\begin{equation}
\mathrm{nDCG@}K=\frac{\mathrm{DCG@}K}{\mathrm{IDCG@}K},
\end{equation}
where $\mathrm{IDCG@}K$ is the ideal DCG achieved by ranking all positives first; and the area under the free-response ROC curve (\textbf{normalized AUFROC}), computed over the top-$K$ ranked candidates.
We sort instances by increasing abnormality score and sweep the cutoff $k=1,\dots,K$, treating the top-$k$ as predicted positives. At each cutoff,
\begin{equation}
\mathrm{TPR}(k)=\frac{\mathrm{TP}(k)}{T_{\max}},\qquad 
\mathrm{FPI}(k)=\frac{\mathrm{FP}(k)}{K},
\end{equation}
where $\mathrm{TP}(k)$ and $\mathrm{FP}(k)$ are the numbers of true/false positives within the top-$k$ list, $T_{\max}$ is the maximum number of positives for that witness rate, and $K$ is the evaluation list size (here $K{=}400$, or $K{=}\min(400,N)$ if fewer predictions are available). The normalized AUFROC is then computed by trapezoidal integration of $\mathrm{TPR}$ as a function of $\mathrm{FPI}$ over $k\le K$.

\medskip
\noindent\textbf{Oral cancer dataset (expert review of top candidates).}
For the oral cancer dataset, cell-level ground truth is unavailable; therefore, we evaluate performance via blinded expert review of retrieved candidates. For each test slide, we rank all extracted $192\times192$ cell patches by the model score and present only the top-100 candidates. To mitigate positional and contextual bias, we anonymize case identifiers, randomize the order of slides (normal vs.\ malignant) presented to the experts, and shuffle the order of the 100 patches within each slide (i.e., patches are not shown in model-ranked order). Two experts (a cytologist and a cytotechnologist) independently review the candidate sets and click on patches they deem suspicious/malignant. We summarize results as the number of expert-selected patches per set of 100 candidates and report these outcomes for malignant test slides.

\subsection{Implementation Details}
\label{sec:impl_details}

\subsubsection{Bone marrow dataset}
Across all methods, we use a ResNet18 backbone and train/evaluate at the instance (cell-patch) level. Input patches are processed with standard resizing/cropping and ImageNet-style normalization where applicable.

\medskip
\noindent\textbf{FS-SIL.}
We train a ResNet18 classifier with cross-entropy on patch-level ground-truth labels using SGD (learning rate $10^{-3}$, batch size $64$). Training uses the same weak color/geometry augmentations as the other patch-based baselines (e.g., \texttt{RGBShift} and \texttt{RandomResizedCrop}), followed by standard normalization.

\medskip
\noindent\textbf{WS-SIL.}
We train the same ResNet18 architecture under slide-to-patch label inheritance using the same optimizer family and data pipeline as FS-SIL (SGD, learning rate $10^{-3}$, batch size $64$), but with inherited (noisy) instance labels.

\medskip
\noindent\textbf{DSVDD (task-specific adaptation).}
We follow an autoencoder-pretraining pipeline before one-class training. The autoencoder is trained on normal patches for $100$ epochs using Adam (learning rate $10^{-4}$, batch size $64$). We then initialize the DSVDD center from pretrained features and apply an elementwise center clamp with $\varepsilon=0.1$ to avoid near-zero collapse. DSVDD training is performed for $200$ epochs with Adam (learning rate $10^{-4}$, batch size $64$) and $\ell_2$ weight decay regularization. At test time, we use mild test-time augmentation (horizontal flip and small rotations) and aggregate view scores using a view-aware rule (blending the original-view score with the maximum-view score; blend weight $0.35$). Unless stated otherwise, we report ensemble results by averaging scores across multiple random seeds.

\medskip
\noindent\textbf{DROC (cytology adaptation).}
We train a SimCLR-style encoder with a 2-layer projection head (projection dimension $256$) for $100$ epochs using Adam (learning rate $10^{-3}$, batch size $64$) and temperature $\tau=2$. Pseudo-abnormals are generated via strong augmentations (our cytology adaptation), while both normal and pseudo-abnormal views receive weak SimCLR-style augmentations (including \texttt{RGBShift} and \texttt{RandomResizedCrop} to $224\times224$, followed by normalization). After contrastive pretraining, we discard the projection head, extract encoder embeddings for the normal training set, and fit a one-class SVM with RBF kernel ($\nu=0.1$, $\gamma=\texttt{auto}$). Test anomaly scores are obtained from the SVM decision function on encoder embeddings. For efficiency, we reuse the pretrained encoder + SVM from higher witness-rate settings when running the lowest witness-rate variants.

\medskip
\noindent\textbf{ItS2CLR.}
We use the official ItS2CLR training recipe with a ResNet18 encoder (output dimension $256$), batch size $512$, and $100$ epochs of training using SGD (learning rate $0.01$, weight decay $10^{-4}$) with cosine annealing. The supervised contrastive temperature is set to $0.07$. The self-paced curriculum uses a warm-up of $10$ epochs; after warm-up, pseudo-positives and pseudo-negatives are selected from slide-positive bags using the aggregator’s instance-confidence scores by taking the top-$r\%$ and bottom-$r\%$ instances, respectively, where $r$ is gradually increased over training (self-paced schedule). MIL refinement is run periodically (every $10$ epochs), and the MIL module is trained with Adam (learning rate $2\times10^{-4}$, weight decay $10^{-4}$) for up to $200$ epochs per refinement stage.

\subsubsection{Oral cancer dataset}

All oral-cancer experiments use a ResNet18 backbone and operate on cell-centered RGB patches of size $192\times192$ pixels. Where required by the training pipeline, patches are resized to $224\times224$.

\medskip
\noindent\textbf{WS-SIL (label inheritance).}
We train a ResNet18 patch classifier using slide-to-patch label inheritance with SGD (learning rate $10^{-3}$, batch size $256$) for 60 epochs, and save checkpoints at epochs 30 and 60.

\medskip
\noindent\textbf{DSVDD.}
We train DSVDD using an autoencoder-pretraining stage followed by one-class training with Adam (learning rate $10^{-4}$, batch size $512$) and latent dimension $d{=}32$; both stages are run for 150 epochs. For the bone marrow experiments, the effective normal training set used by the one-class methods is comparatively small (only 9185 patches) and the ultra-low witness-rate regimes can be sensitive to initialization and minor appearance variations, so we use a seed ensemble and mild test-time augmentation (TTA) there to reduce variance and improve robustness of the anomaly ranking. In contrast, for the oral cytology dataset we train on a substantially larger pool of slide-negative patches (on the order of $\sim$640k), which stabilizes representation learning and makes the additional computational cost of multi-seed training and multi-view inference less beneficial; we therefore keep DSVDD training and inference deterministic and lightweight for large-scale slide-wise scoring.

\medskip
\noindent\textbf{DROC}
We train a SimCLR-style ResNet18 encoder with a 2-layer projection head (projection dimension 256) for 100 epochs using Adam (learning rate $10^{-3}$, batch size $64$, temperature $\tau{=}2$). We generate pseudo-abnormal variants using strong augmentations (our oral-cytology adaptation: \textsc{CenterCropResize}, \textsc{GridDistortion}, and \textsc{ColorJitter}). After pretraining, we discard the projection head, extract encoder embeddings for the healthy training patches, and fit a one-class SVM (RBF kernel, $\nu{=}0.1$, $\gamma{=}\texttt{auto}$). Slide-wise rankings are produced by sorting patch scores within each test slide.

\medskip
\noindent\textbf{ItS2CLR.}
We use the same ItS2CLR configuration as in the bone marrow experiments and apply it under slide-level bag supervision on the oral dataset, producing patch scores used for top-$100$ retrieval and expert review.

\section{Results}

\subsection{Bone marrow dataset}
\label{subsec:results_bm}

\noindent\textbf{Primary metric: Recall@400 (top-$K$ retrieval).}
Our evaluation targets a clinically motivated workflow where a cytologist reviews only a small shortlist of top-ranked candidates rather than the full cell population. We therefore emphasize \textbf{Recall@400} as the primary measure of effectiveness. For the qualitative evaluation in this study, we treat the top-$K$ predictions as an \emph{unordered} set to avoid positional bias toward the highest-ranked items and to assess how effectively the model surfaces suspicious cells within the retrieved set. In practical use, however, these candidates would naturally be presented in ranked order according to the model score.

\begin{figure}[pos=t]
    \centering
    \includegraphics[width=\linewidth]{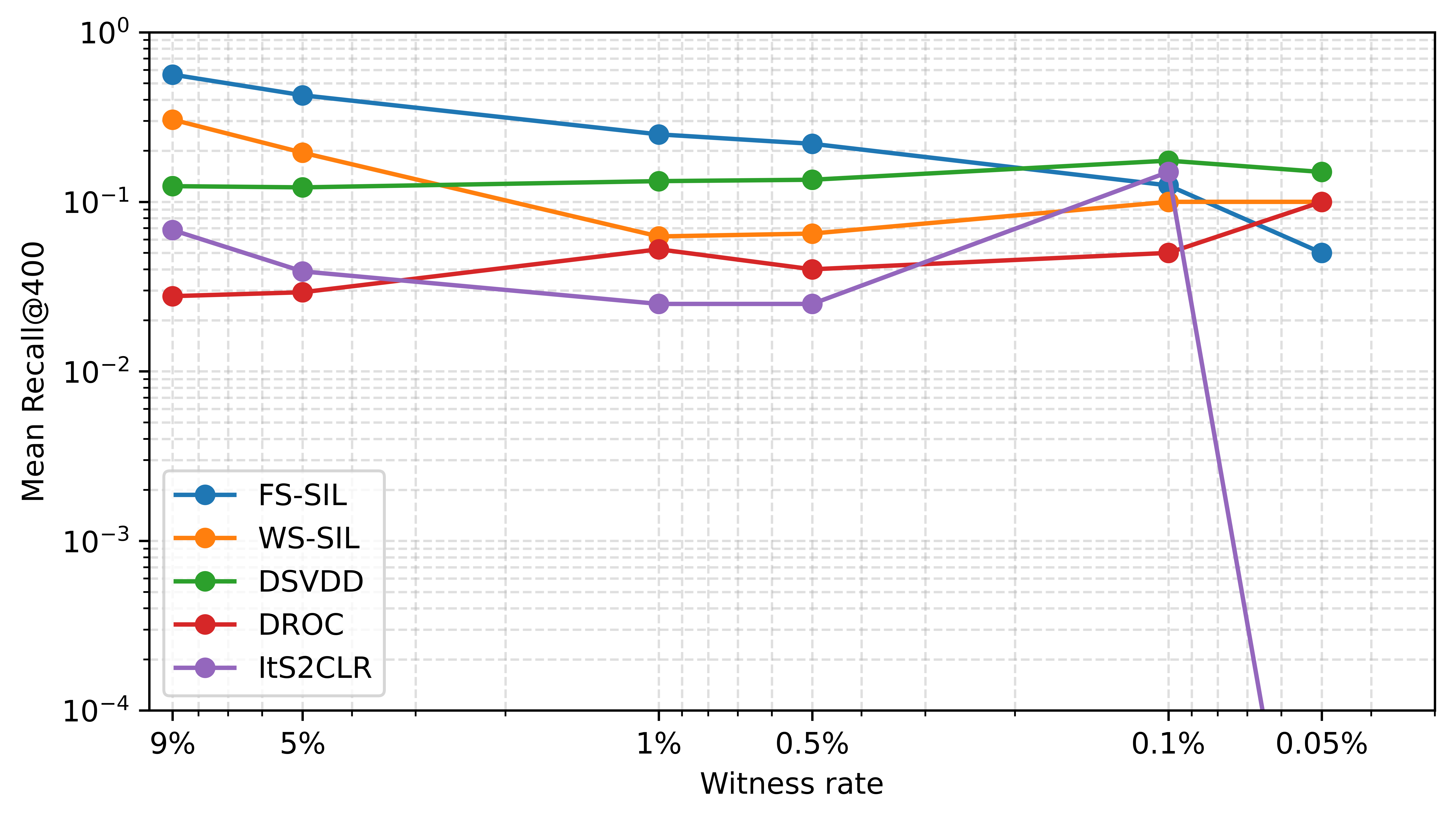}
    \caption{Mean Recall@400 on the bone marrow dataset across witness rates (log-log scale).}
    \label{fig:bm_recall}
\end{figure}

As shown in Fig.~\ref{fig:bm_recall}, performance degrades with decreasing witness rate (WR) for methods that rely on weak or noisy positive supervision, whereas one-class approaches remain comparatively stable. In the very-low WR regime ($\mathrm{WR}\leq 1\%$), \textbf{DSVDD} consistently provides the strongest retrieval among the weak/one-class baselines, highlighting the benefit of explicitly learning a compact representation of normality when malignant cells are exceedingly rare. While the fully supervised upper bound (FS-SIL) remains strongest when exhaustive instance labels are available, our main focus is the behavior of weak/one-class methods under extreme rarity.

\vspace{1mm}
\noindent To make these trends directly interpretable in terms of retrieved instances, Table~\ref{tab:tp400_counts} reports the mean number of true positives retrieved in the top-$400$ (TP@400).

\begin{table}[t]
\caption{Mean true positives in the top-400 (TP@400) on the bone marrow dataset (mean$\pm$std). Bold values indicate the top values across witness rates. The fully supervised FS-SIL is not considered a realistic option (and is therefore grayed out) since it relies on instance level annotations, and is included only to indicate an expected upper bound of performance.}
\label{tab:tp400_counts}
\centering
\small
\setlength{\tabcolsep}{3pt}
\begin{tabular}{@{}lccccc@{}}
\toprule
WR & \textcolor{gray}{FS-SIL} & WS-SIL & DSVDD & DROC & ItS2CLR \\
\midrule
9\%    & \textcolor{gray}{223.0$\pm$0.0} & \textbf{121.0}$\pm$0.0 & 49.0$\pm$0.0 & 11.0$\pm$0.0 & 27.0$\pm$0.0 \\
5\%    & \textcolor{gray}{84.2$\pm$3.6}  & \textbf{38.6}$\pm$2.6  & 24.1$\pm$2.5 & 5.8$\pm$1.3  & 7.7$\pm$4.2  \\
1\%    & \textcolor{gray}{10.0$\pm$2.3}  & 2.5$\pm$2.1   & \textbf{5.3}$\pm$1.6 & 2.1$\pm$1.3  & 1.0$\pm$0.9  \\
0.5\%  & \textcolor{gray}{4.4$\pm$2.2}   & 1.3$\pm$1.0   & \textbf{2.7}$\pm$1.5 & 0.8$\pm$1.0  & 0.5$\pm$0.5  \\
0.1\%  & \textcolor{gray}{0.5$\pm$0.7}   & 0.4$\pm$0.5   & \textbf{0.7}$\pm$0.6 & 0.2$\pm$0.4  & 0.6$\pm$0.7  \\
0.05\% & \textcolor{gray}{0.1$\pm$0.3}   & 0.2$\pm$0.4   & \textbf{0.3}$\pm$0.5 & 0.2$\pm$0.6  & 0.0$\pm$0.0  \\
\bottomrule
\end{tabular}
\end{table}

\begin{figure}[pos=h]
    \centering
    \includegraphics[width=\columnwidth]{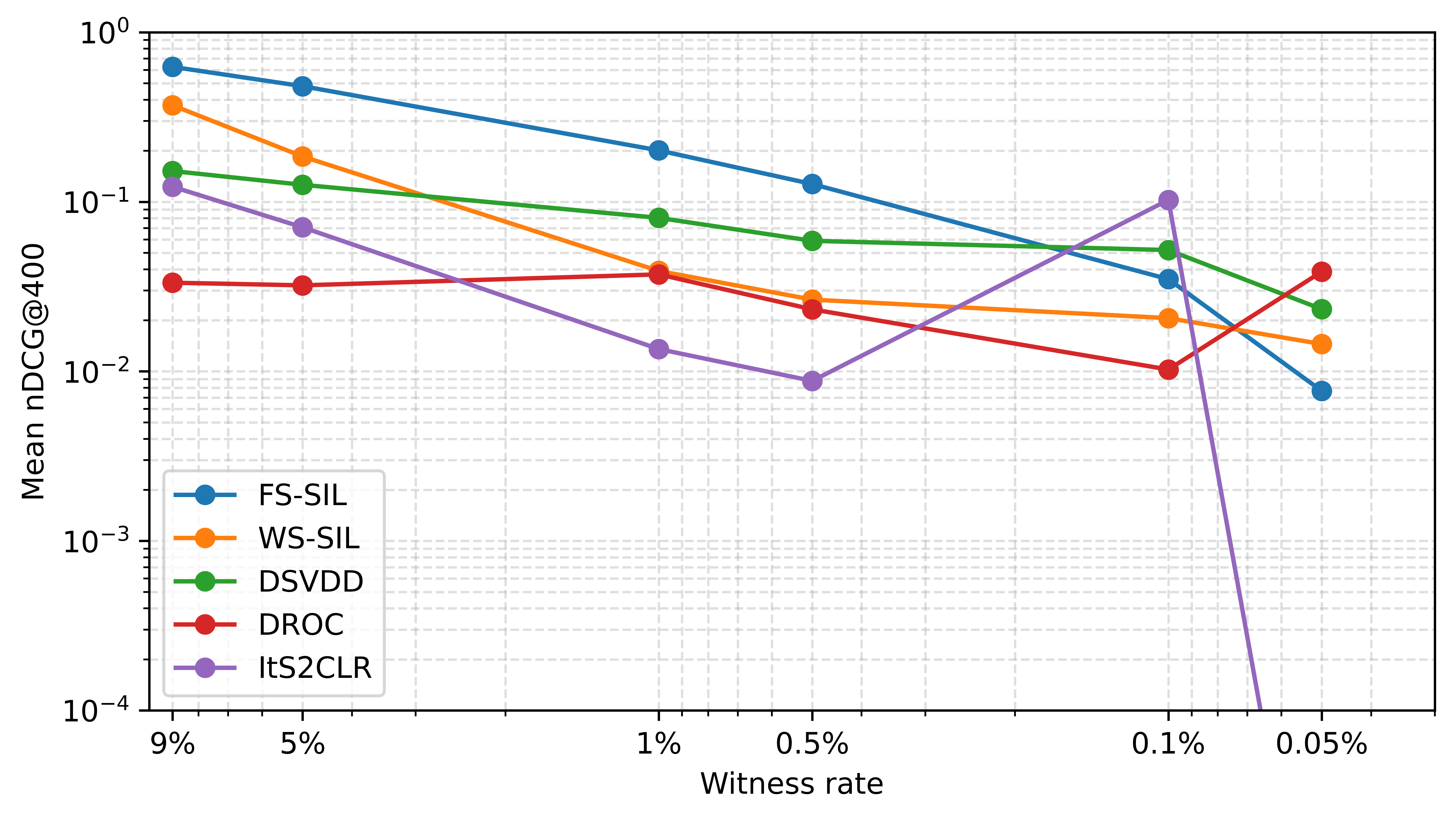}
    \caption{Mean nDCG@400 on the bone marrow dataset across witness rates (log-log scale).}
    \label{fig:bm_ndcg}
\end{figure}
\begin{figure}[pos=h]
    \centering
    \includegraphics[width=\columnwidth]{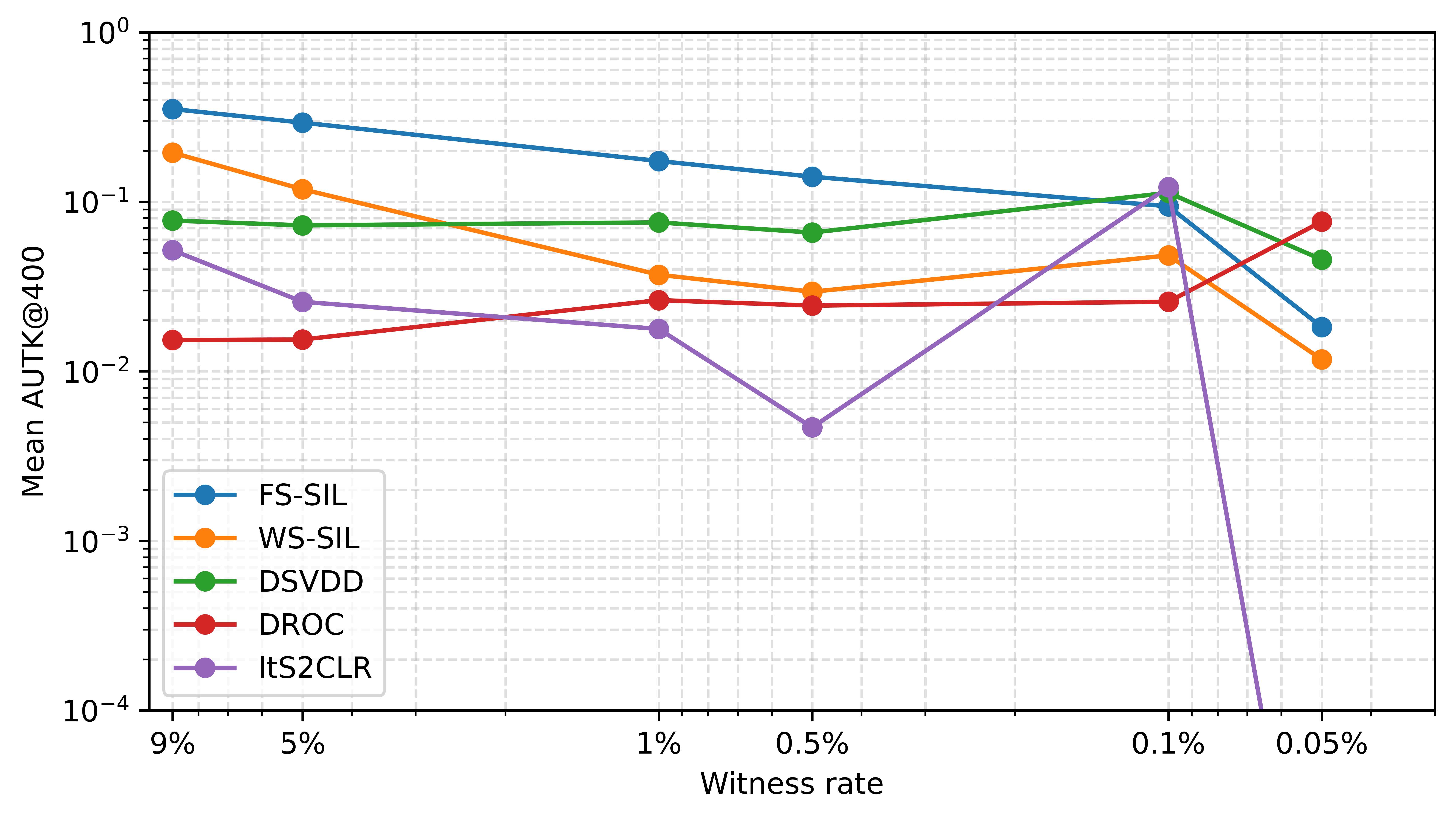}
    \caption{Mean AUTK@400 on the bone marrow dataset across witness rates (log-log scale).}
    \label{fig:bm_autk}
\end{figure}
\begin{figure}[pos=h]
    \centering
    \includegraphics[width=\columnwidth]{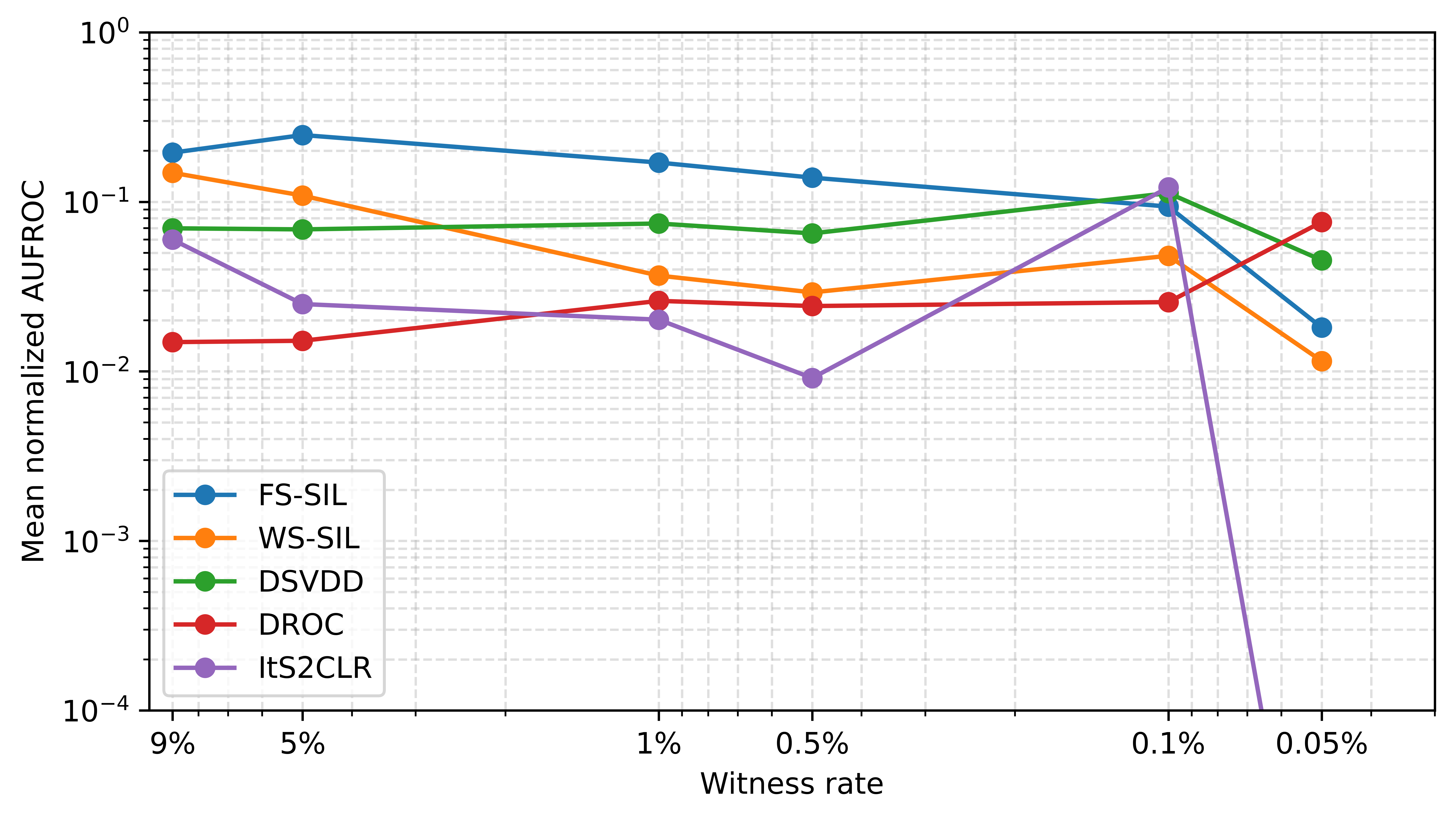}
    \caption{Mean normalized AUFROC on the bone marrow dataset across witness rates (log-log scale).}
    \label{fig:bm_froc}
\end{figure}

\noindent\textbf{Secondary metrics: nDCG@400, AUTK@400, and normalized AUFROC@400.}
To analyze \emph{where} true positives appear within the top-$400$ (earlier vs.\ later ranks), we additionally report nDCG@400 and AUTK@400 (Figs.~\ref{fig:bm_ndcg} and \ref{fig:bm_autk}). We further report the \emph{normalized} AUFROC@400 (Fig.~\ref{fig:bm_froc}), which summarizes the retrieval trade-off between true-positive discovery and false positives within the top-$K$ list.

These ranking-sensitive measures broadly mirror the recall trend: performance typically decreases as WR drops for methods relying on weak positive supervision, while one-class approaches remain comparatively stable. In particular, AUTK and normalized AUFROC highlight that \textbf{DSVDD} maintains a favorable early-retrieval trade-off across low-WR regimes, whereas WS-SIL and FS-SIL tend to degrade as the shortlist becomes dominated by false positives. We also observe an apparent jump for ItS2CLR at $\mathrm{WR}=0.1\%$, followed by a collapse at $\mathrm{WR}=0.05\%$ (consistent with its failure to identify predicted-positive bags in this extreme setting). We do not over-interpret such behavior at ultra-low WRs, where results can be dominated by stochastic effects (e.g., the particular abnormal instances sampled and their visual similarity to normal cells). Overall, our primary conclusion is stable: in the low-WR regime ($\mathrm{WR}\leq 1\%$), \textbf{DSVDD} provides the most consistent instance retrieval among the weak/one-class methods, which is the main focus of this study.

\noindent\textbf{High-WR regime (5\% and 9\%).}
At higher witness rates, WS-SIL and ItS2CLR benefit from increased exposure to abnormal instances via weak supervision and therefore improve their retrieval performance. However, our central objective is to assess one-class representation learning for the realistic screening regime where abnormal cells are exceedingly rare ($\mathrm{WR}\leq 1\%$), and under this condition DSVDD is the strongest and most stable weak/one-class baseline.

\begin{figure}[pos=h]
    \centering
    \includegraphics[width=\linewidth]{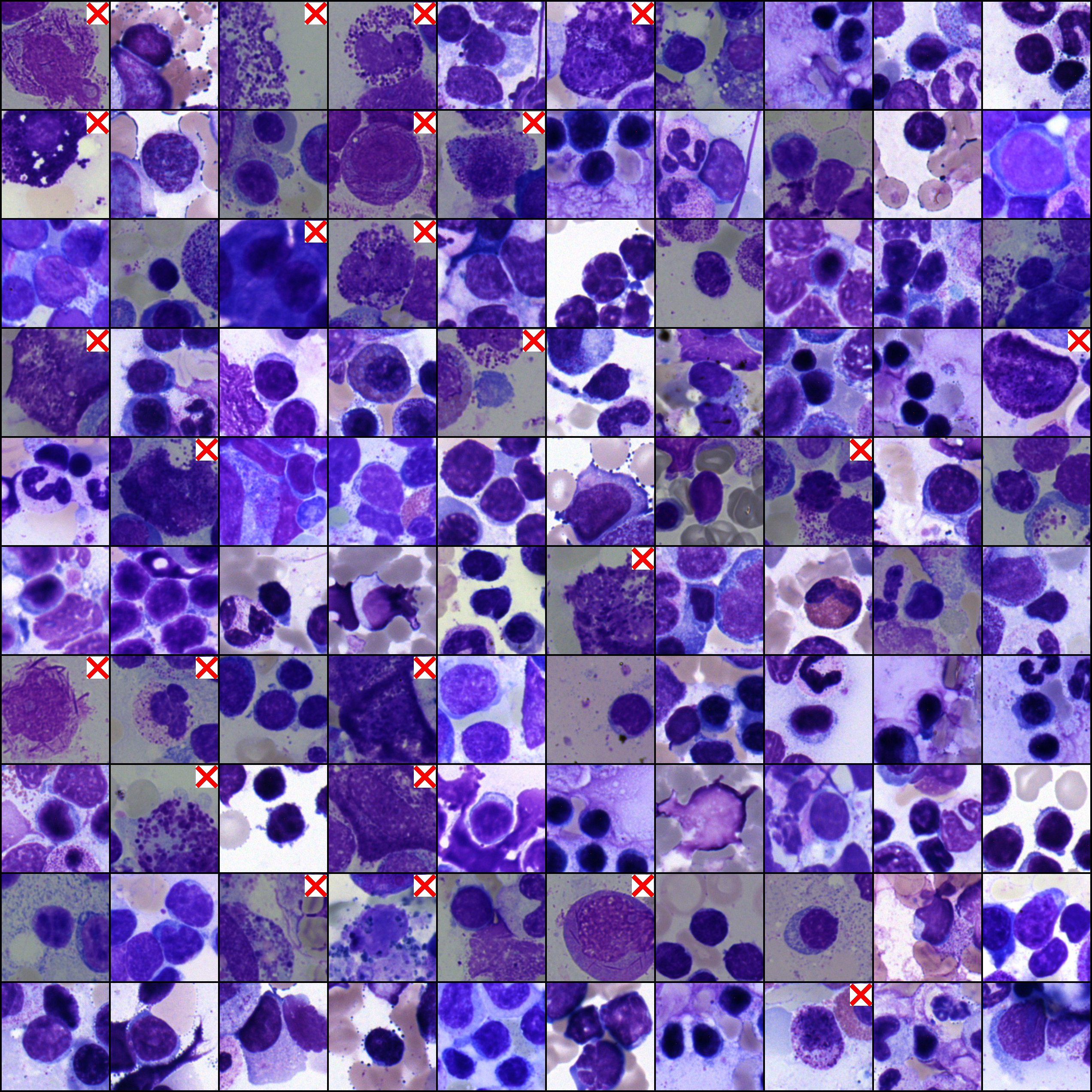}
    \caption{Example mosaic for the bone marrow dataset at 10\% WR. Shown are the top 100 patches (unordered) using DSVDD anomaly scores and arranged in a $10\times10$ grid. A small white square in
    the top-right corner of a patch indicates that the underlying instance is annotated as non-LYT (i.e.\ abnormal or suspicious); within this box, `$\times$' marks such non-LYT cells. Patches without a white box
    correspond to LYT (normal) cells. For this mosaic, 25 of the top 100
    predictions were true positives (non-LYT).}
    \label{fig:bm_mosaic}
\end{figure}

\noindent\textbf{Example grid for expert review}
Figure~\ref{fig:bm_mosaic} illustrates the intended expert-facing output of these frameworks: for each slide (or merged evaluation pool), the model assigns an abnormality score to every detected cell and returns a small set of the highest-scoring candidates for rapid inspection. As a concrete demonstration, we show a sample $10\times10$ grid assembled from the bone marrow dataset by arranging the top-$100$ ranked patches. This visualization provides a compact, human-readable summary that a cytologist can review quickly. In whole-slide cytology, where a single sample can contain on the order of $10^5$ cells, presenting only the top-$K$ candidates can substantially reduce workload while preserving transparency, the expert directly inspects the specific cells that the model considers most suspicious.

\subsection{Oral cancer dataset}
\label{subsec:results_oral}

Unlike the bone marrow dataset, the oral cancer WSCI dataset provides only slide-level ground truth (healthy vs.\ malignant) and does not include exhaustive cell-level annotations indicating which nuclei-centered patches are truly malignant. Moreover, a single oral slide can contain on the order of $10^5$ detected nuclei (Table~\ref{tab:nuclei_counts}), making exhaustive expert verification infeasible even for one slide. For these reasons, we do not report rank-based retrieval metrics (e.g., Recall@$K$, nDCG@$K$, AUTK@$K$, normalized AUFROC@$K$) on this dataset.

Instead, we evaluate clinical utility via blinded expert review of the model-selected top-100 candidates per malignant test slide, presented as an unordered $10\times10$ grid (randomized order) to minimize positional bias. For each of the four malignant test slides, we rank all extracted $192\times192$ patches by the method’s abnormality score and present only the top-100 candidates to two experts (a cytologist and a cytotechnologist). The experts independently inspect the 100-patch set and mark patches they deem suspicious/malignant. Table~\ref{tab:cyto_evals} summarizes the number of expert-marked patches among the top-100 candidates for each method and malignant slide (reported as a tuple of counts from the two experts). Overall, DSVDD yields the most consistent expert-confirmed relevance across slides—showing higher total (and average) expert markings on multiple malignant cases compared to the other methods in Table~\ref{tab:cyto_evals}.

We also note that, even on healthy test slides, the models frequently surface patches that trained experts consider visually concerning (e.g., suspicious-looking but benign nuclei, poor/degenerate morphology, or cells in slides with a strong inflammatory component). This suggests that the learned ``abnormality'' scores capture broader deviations from typical healthy appearance, not only overt malignancy.

\begin{table}[t]
\centering
\caption{Expert assessment of top-100 patches on the four malignant oral-cancer test slides. Tuples report the number of patches marked as suspicious by the cytologist and cytotechnologist (in this order). Bold indicates the slides visualized in Fig.~\ref{fig:oral_mosaics}.}
\label{tab:cyto_evals}
\resizebox{\columnwidth}{!}{%
\begin{tabular}{@{}lcccc@{}}
\toprule
\textbf{Type} & WS-SIL & DSVDD & DROC & ItS2CLR \\
\midrule
Malignant 1  & (0,0)  & (0,0)  & (0,2)    & \textbf{(1,2)} \\
Malignant 2  & \textbf{(25,93)} & (18,63) & (2,65)  & (2,47) \\
Malignant 3  & (7,46) & \textbf{(9,52)} & (18,19)  & (4,19) \\
Malignant 4  & (8,11) & \textbf{(4,22)} & (9,16)   & (2,10) \\
\bottomrule
\end{tabular}%
}
\end{table}

\begin{figure*}[pos=h]
    \centering
    \begin{subfigure}[t]{0.48\textwidth}
        \centering
        \includegraphics[width=\linewidth]{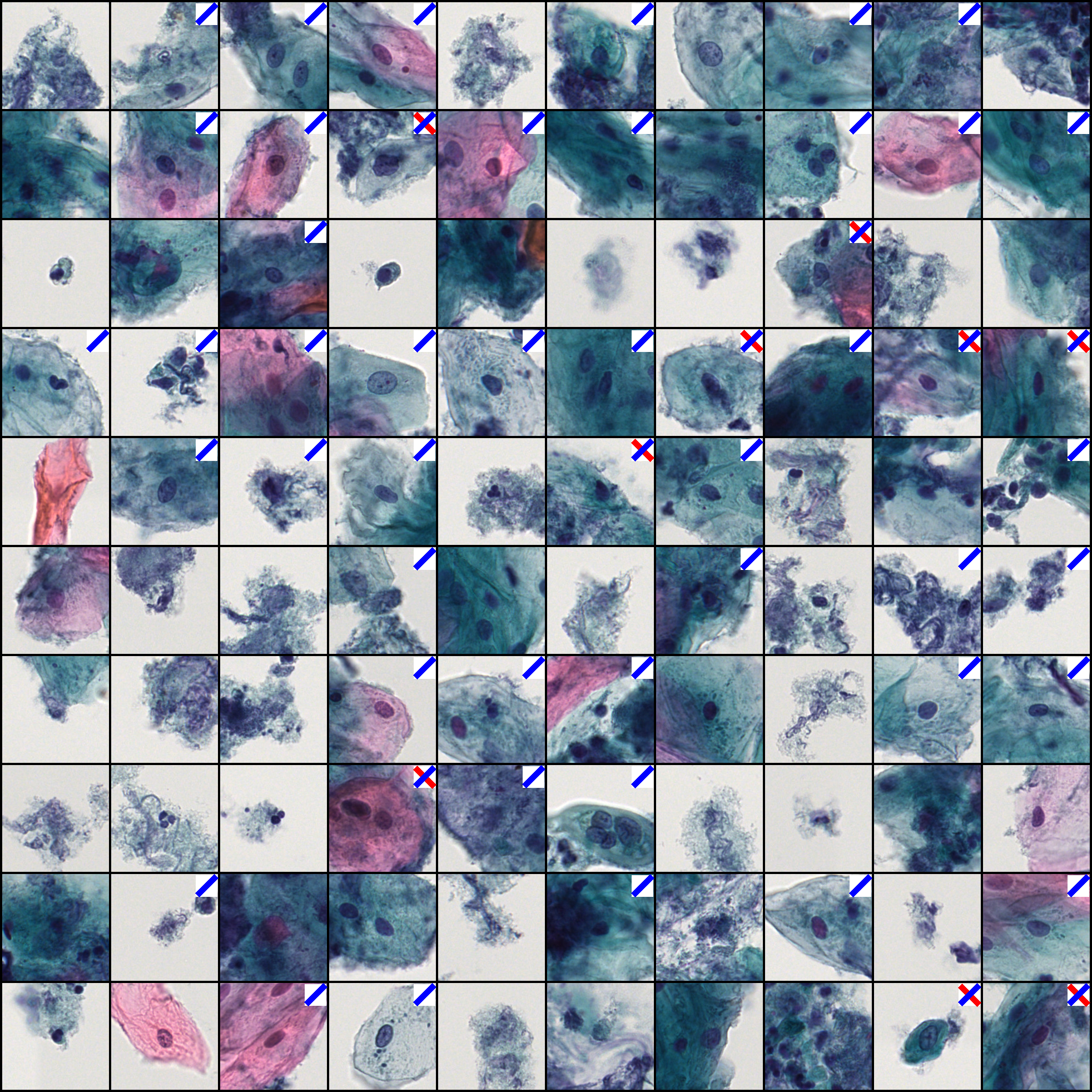}
        \caption{Malignant slide 3: 9 patches were marked by both experts.}
        \label{fig:oral_mosaic_a008}
    \end{subfigure}\hfill
    \begin{subfigure}[t]{0.48\textwidth}
        \centering
        \includegraphics[width=\linewidth]{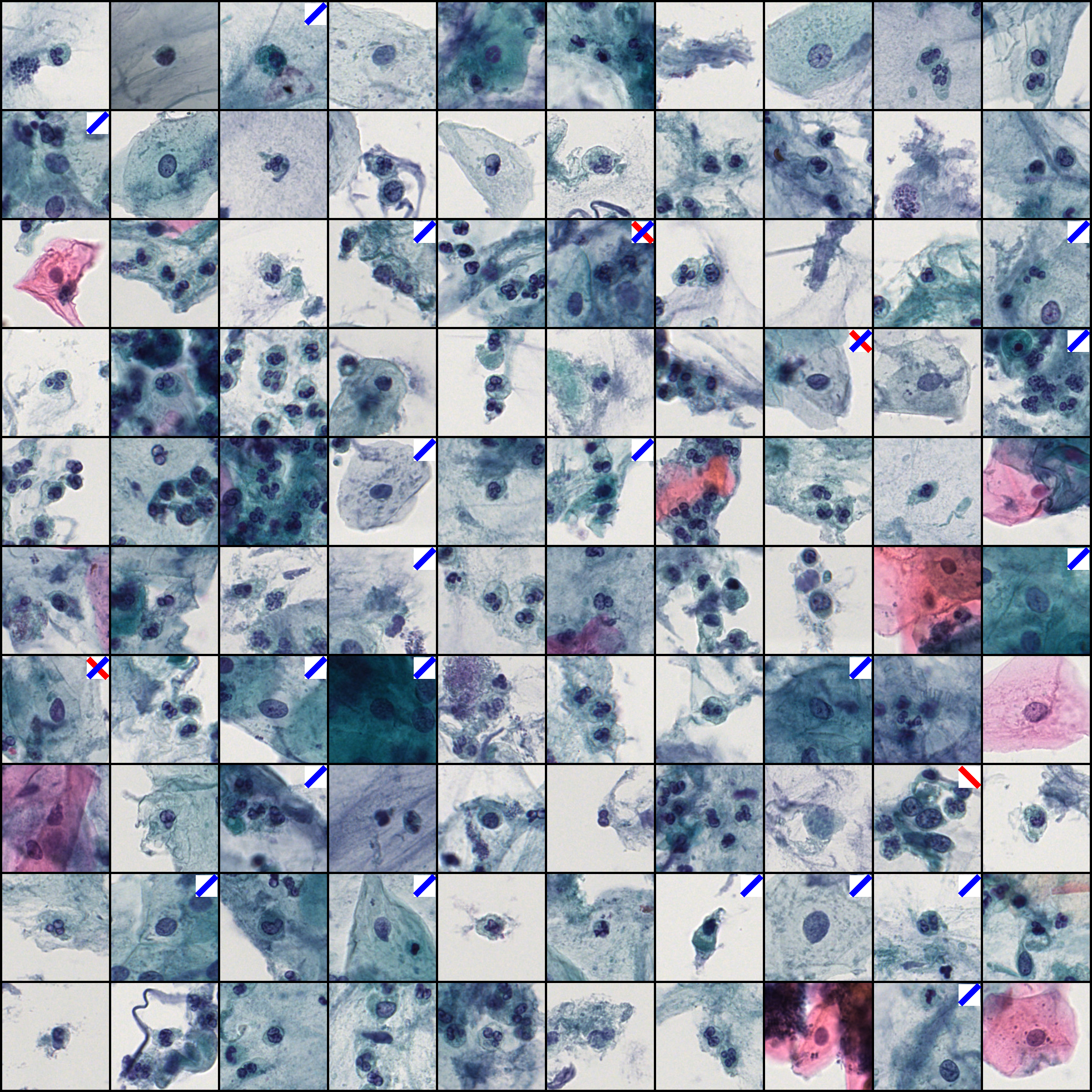}
        \caption{Malignant slide 4: 3 patches were marked by both experts.}
        \label{fig:oral_mosaic_a006}
    \end{subfigure}
    \caption{DSVDD-selected top-100 expert-review grids for two malignant test slides (shown in randomized order). For each slide, we show the unordered top-100 patches (arranged as a $10\times10$ grid). A white square in the top-right corner of a patch indicates that at least one expert marked it as suspicious. Within the square, a \textcolor{red}{red} backslash `\textcolor{red}{\bf\textbackslash}' denotes a mark by one expert, an \textcolor{blue}{blue} forward slash `\textcolor{blue}{\bf /}' denotes a mark by the other, and a full cross indicates agreement (both experts).}
    \label{fig:oral_mosaics}
\end{figure*}

To illustrate how our one-class framework could support cytologists on oral cytology WSCIs, we visualize the \emph{top-100} DSVDD candidates for two malignant test slides as unordered $10\times10$ grids (Figure \ref{fig:oral_mosaics}). Since the goal is to provide a compact set of suspicious patches for rapid inspection (rather than to force attention to a specific rank order), we present these candidates in randomized order within each grid and overlay the independent markings from a cytologist and a cytotechnologist.

\section{Discussion}

Detecting rare malignant cells in whole-slide cytology is challenging because positives are extremely low in number, morphology is heterogeneous, and dense instance-level annotation is impractical. Our results show that one-class representation learning, trained only on slide-negative patches, is a strong alternative to weakly supervised learning when witness rates are very low and when instance-level localization (rather than only slide-level classification) is the primary goal.

\medskip
Our findings across the bone marrow and oral-cancer datasets suggest the following:

\noindent\textbf{One-class learning is robust under extreme rarity.}
In the low-WR regime ($\mathrm{WR}\leq 1\%$), DSVDD consistently produced stronger and more stable retrieval  behavior than weakly supervised baselines, as reflected by Recall@400 and the ranking-sensitive metrics (nDCG@400, AUTK@400, and normalized AUFROC@400). This supports the premise that learning a compact normality manifold can be more reliable than learning from weak slide-level supervision when abnormal instances are vanishingly rare.

\noindent\textbf{Seed ensembling + mild TTA stabilize anomaly ranking in the small-data regime.}
For bone marrow, the effective normal training set for one-class learning is comparatively small (9,185 patches) and the lowest-WR settings are sensitive to initialization and minor appearance variation; multi-seed ensembling and mild test-time augmentation reduce variance in the anomaly ranking. In contrast, the oral-cancer setting offers a much larger pool of slide-negative patches (on the order of $\sim$640k), which stabilizes representation learning and makes multi-seed/multi-view inference less beneficial relative to its added computational cost.

\noindent\textbf{DROC is competitive but augmentation design matters.}
DROC provides a principled way to couple self-supervised contrastive learning with one-class detection via distribution augmentation. Its performance depends on how well pseudo-abnormal augmentations reflect meaningful departures from normal cytology; overly aggressive or misaligned transforms can reduce separability between true abnormalities and hard normal variants.

\noindent\textbf{Expert-facing utility on oral cancer can be assessed without exhaustive labels.}
Because cell-level ground truth is unavailable for oral slides, we evaluated clinical utility via blinded expert review of the model-selected top-100 candidates per malignant slide (presented as an unordered grid to reduce positional bias). In this setting, DSVDD yielded higher expert-marked counts on average across malignant slides than the other weak/one-class baselines, suggesting that one-class scoring can produce practically useful candidate sets even when quantitative retrieval metrics cannot be computed.

\noindent\textbf{Weak supervision alone is brittle at ultra-low WR.}
Slide-to-patch label inheritance (WS-SIL) and bag-supervised selection methods can be dominated by label noise and spurious slide-level cues when only a tiny fraction of instances are truly abnormal. This can lead to unstable instance rankings and reduced early retrieval quality, particularly as WR decreases.

\noindent\textbf{Choosing $K$ and handling high-burden slides.}
The review budget $K$ is clinically consequential: while we use $K{=}400$ to reflect a practical inspection limit, advanced-stage malignant slides may contain far more than $K$ malignant cells, so any fixed-$K$ protocol caps achievable recall by design. A practical extension is to complement top-$K$ visualization with an adaptive rule that thresholds the anomaly score to estimate how many instances on a slide are likely malignant, while still displaying only the top-$K$ most suspicious patches for rapid expert confirmation.

\section{Limitations and future work.}
We highlight key directions to strengthen robustness and clinical utility:

\noindent\textbf{Cytology-aware pseudo-abnormals.} DROC depends on heuristic pseudo-abnormal augmentations; future work should design stain-/morphology-aware perturbations that better reflect clinically meaningful deviations.

\noindent\textbf{Domain shift + hard normals.} OCC methods can degrade under scanner/stain/focus/debris variation and in the presence of ``suspicious but benign'' nuclei; domain-invariant pretraining, normalization, and explicit handling of hard normal variants are important.

\noindent\textbf{Oral-cancer evaluation scale.} Oral-cancer results rely on blinded expert review rather than exhaustive instance labels; larger cohorts, prospective studies, and structured partial verification (including hard negatives) would enable more quantitative assessment.

\noindent\textbf{Choosing $K$ on high-burden slides.} A fixed review budget caps achievable recall when malignant burden exceeds $K$; a practical extension is to combine top-$K$ visualization with adaptive score-thresholding to estimate burden while still showing only the most suspicious patches.

\noindent\textbf{Expert-in-the-loop refinement.} Iterative top-$K$ review with feedback from multiple experts (active learning) can refine rankings over rounds without requiring dense slide-wide annotation.

\section{Conclusion}

We study rare malignant-cell localization in whole-slide cytology under an expert-facing top-$K$ review setting. On the fully annotated bone-marrow benchmark, one-class learning with DSVDD is the most consistent method in the low-witness regime ($\mathrm{WR}\leq 1\%$), improving Recall@400 and ranking-sensitive metrics (AUTK@400, and normalized AUFROC@400) over weakly supervised baselines. On oral cancer, where cell-level ground truth is unavailable, blinded expert review suggests that one-class scoring can surface clinically relevant candidates for inspection. Importantly, the output of all methods is naturally interpretable (in a bottom-up approach) as a small set of visual candidates (expert-review grids), supporting practical triage without requiring dense annotations. Future work includes cytology-aware pseudo-abnormal generation, reducing domain shift, and developing fine-grained instance-level explanations to further strengthen clinical trust.

\section*{Acknowledgements}

\noindent\textbf{Expert support.} We sincerely thank the participating cytologist and cytotechnologist for their invaluable support, including multiple rounds of expert review and annotation of the oral-cancer test slides.

\noindent\textbf{Funding.} This work is supported by Sweden’s Innovation Agency (VINNOVA) projects 2017-02447, 2020-03611, 2021-01420, Swedish Cancer Society (Cancerfonden) projects 22 2353 Pj and 222357 Pj., and the Centre for Interdisciplinary Mathematics (CIM), Uppsala University.

\noindent\textbf{Figure asset credits.} Icons used in the TikZ pipeline illustration are credited as follows: “computer screen” by Vectors Point (The Noun Project, CC BY 3.0); “Doctor” by Hanbai (The Noun Project, CC BY 3.0); “slide scanner” by Sukjun Kim (The Noun Project, CC BY 3.0); “open mouth” icon designed by Smashicons / Freepik (Asset ID: 1471596); and “blood” (test tube) icon designed by Smashicons / Freepik (Asset ID: 1856969).

\noindent\textbf{Declaration of generative AI and AI-assisted technologies in the manuscript preparation process.}
During the preparation of this work the author(s) used ChatGPT in order to refine or edit manuscript language. After using this tool/service, the author(s) reviewed and edited the content as needed and take(s) full responsibility for the content of the published article.

\subsection*{Ethics statement}
This retrospective study used already collected and fully de-identified data; no new human data were collected for this work. Single-cell image patches do not allow patient-specific identification or tracking. The study was conducted in accordance with the Declaration of Helsinki and the Swedish Ethical Review Act. Use of the oral cytology dataset is covered by ethical approvals from the Swedish Ethical Review Authority (permit 2015/1213-31, with addenda 2016/1263-32 and 2017/1535-32, and permit 2019-00349). All included subjects received written and oral participant information and provided informed consent. The bone-marrow dataset is based on publicly available, de-identified data; its use was approved by the MLL Munich Leukemia Laboratory internal institutional review board, and all patients provided written informed consent for use of clinical data in accordance with the Declaration of Helsinki. No animal experiments were performed.

\bibliographystyle{unsrtnat}
\bibliography{cas-refs}

\end{document}